\documentclass[runningheads]{llncs}
\usepackage{graphicx}
\usepackage{xcolor}
\usepackage{amsmath}
\usepackage{datetime2}
\usepackage{multirow}
\usepackage{enumitem}

\begin{document}
\title{Making sense of violence risk predictions using clinical notes}
\author{Pablo Mosteiro\inst{1} \and
Emil Rijcken\inst{1,2} \and
Kalliopi Zervanou\inst{2} \and
Uzay Kaymak\inst{2} \and
Floortje Scheepers\inst{3} \and
Marco Spruit\inst{1}}
\authorrunning{P. Mosteiro et al.}
\institute{Utrecht University, Utrecht, the Netherlands\\
\email{\{p.mosteiro,m.r.spruit\}@uu.nl} \and
Eindhoven University of Technology, Eindhoven, the Netherlands\\
\email{\{e.f.g.rijcken,k.zervanou,u.kaymak\}@tue.nl} \and
University Medical Center Utrecht, Utrecht, the Netherlands\\
\email{f.e.scheepers-2@umcutrecht.nl}}
\maketitle              % typeset the header of the contribution
\begin{abstract}
  Violence risk assessment in psychiatric institutions enables interventions to avoid violence incidents. Clinical notes written by practitioners and available in electronic health records (EHR) are valuable resources that are seldom used to their full potential. Previous studies have attempted to assess violence risk in psychiatric patients using such notes, with acceptable performance. However, they do not explain \emph{why} classification works and how it can be improved. We explore two methods to better understand the quality of a classifier in the context of clinical note analysis: random forests using topic models, and choice of evaluation metric. These methods allow us to understand both our data and our methodology more profoundly, setting up the groundwork to work on improved models that build upon this understanding. This is particularly important when it comes to the generalizability of evaluated classifiers to new data, a trustworthiness problem that is of great interest due to the increased availability of new data in electronic format.

\end{abstract}
\section*{Notice}
This is a pre-print of the following work: Mosteiro, P., Rijcken, E., Zervanou, K., Kaymak, U., Scheepers, F., Spruit, M. (2020). Making Sense of Violence Risk Predictions Using Clinical Notes. In: Huang, Z., Siuly, S., Wang, H., Zhou, R., Zhang, Y. (eds) Health Information Science. HIS 2020. Lecture Notes in Computer Science(), vol 12435. Springer, Cham. The final authenticated version is available online at: \url{https://doi.org/10.1007/978-3-030-61951-0\_1}

\section{Introduction}
Two thirds of mental health professionals working in Dutch clinical psychiatry institutions report having been a victim of at least one physical violence incident in their careers~\cite{vanleeuwen}. These incidents can have a strong psychological effect on nurses~\cite{inoue}, as well as economical consequences~\cite{nijman}. Multiple approaches have been proposed to predict and avoid violence incidents in the international community, with some adoption in practice~\cite{singh}. However, few of these methods leverage the unstructured textual data contained in patients' Electronic Health Records (EHR).

Machine learning methods have been successfully applied to psychiatric EHR's to predict readmission~\cite{rumshisky}.
Most current applications of text processing in psychiatric EHR's are for the English language~\cite{kim}.
Building up on promising first attempts to systematically analyse EHR's in Dutch~\cite{vincent5,deduce,vincent6}, the COVIDA project (COmputing VIsits DAta) aims to create a publicly available self-service facility for Natural Language Processing (NLP) of Dutch medical texts.

In order to build a self-service tool, it is essential to dig deep into the machine learning methods employed and build confidence and trust in practictioners. In this work, we investigate the problem of predicting violence incidents using unstructured clinical notes in Dutch and attempt to provide better understanding and interpretable results. For this purpose, we re-implement an SVM document classification approach suggested in~\cite{vincent5,vincent6} on a 35\% bigger dataset; we expand on the text features used by combining unstructured text with existing structured data, such as the patient's age; and we experiment with alternative document representation techniques, such as LDA and word embeddings, using random forest classification in order to also gain better insights in potentially significant features.
We find that the results are promising, though much work remains to be done to achieve acceptable performance for the clinical practice.

\section{Related work}
The analysis of free text in EHR's and the combination of these to structured data using machine learning approaches is gaining an increasing interest as anonymised EHR's become available for research. However, the analysis of clinical free-text data presents numerous challenges due to (i) \emph{highly imbalanced data} with respect to the class of interest~\cite{Rijoetal2015}; (ii) \emph{lack of publicly available data-sets}, limiting research on private institutional data~\cite{Wangetal2018}; and (iii) relatively \emph{small data sizes} compared to the amounts of data currently used in text processing research.  

In the psychiatric domain, structured data such as symptom codes and medication history have been used for the prediction of admissions~\cite{Friedmanetal1983,Lyonsetal1997,Olfsonetal2011}. Studies using structured information in EHR's to predict suicide risk~\cite{Conneretal2012} indicate that information from the unstructured data in clinical texts may provide better insights on risk factors and result in better predictions. Free text in combination with structured EHR variables has been used in suicide~\cite{Cooketal2016} and depression diagnosis~\cite{Huangetal2014} among healthy \textit{vs.} unhealthy individuals. In such approaches, structured variables such as medication history, questionnaires, and demographics are expected to provide enough discriminatory power for the required analysis.
Research approaches focusing on unstructured text from EHR's in mental healthcare are to our knowledge very few; Poulin \textit{et al.}~\cite{Poulinetal2014} attempted to predict suicide risk among veterans and more recently Menger \textit{et al.}~\cite{vincent6} used Dutch clinical text to predict violent incidents from patients in treatment facilities. 

The most popular machine learning methods used for processing free text in EHR data 
are support vector machines (SVM), logistic regression, naive Bayes, and decision trees~\cite{Abbeetal2016,AggarwalZhai2012}. Decision-tree classification is one of the easiest to interpret approaches, because it allows for inspection of the specific feature combination used for the classification. This line of classification approaches has also achieved significant improvements in classification accuracy by growing an ensemble of decision trees (a \emph{random forest}) trained on subsets of the data-set and letting them vote for the most popular class~\cite{breiman}.

\section{Dataset}
  \label{sec:dataset}
  The data used in this study consists of clinical notes written in Dutch by nurses and physicians about patients in the psychiatry ward of the University Medical Center (UMC) Utrecht between 2012-08-01 and 2020-03-01. The 834834 notes available are de-identified for patient privacy using DEDUCE~\cite{deduce}.

  Each patient can be admitted to the psychiatry ward multiple times. In addition, within an admission a patient can be sent to various sub-departments of psychiatry. The time the patient spends in each of the sub-departments is called an \emph{admission period}. In the present study, our datapoints are admission periods. For each admission period, all notes collected between 28 days before and 1 day after the start of the admission period are concatenated and considered as a single \emph{period note}. If a patient is involved in a violence incident between 1 and 28 days after the start of the admission period, the outcome is recorded as \emph{violent} (hereafter also referred to as \emph{positive}). Otherwise, it is recorded as non-violent. Admission periods having period notes with fewer than or equal to 100 words are discarded as was done in previous work~\cite{vincent6,rumshisky}.

  In addition to notes, we employ structured variables collected in various formats by the hospital. These include variables related to:
  \begin{itemize}
  \item Admission periods (\textit{e.g.}, start date and time)
  \item Notes (\textit{e.g.}, date and time of first \& last notes in period)
  \item Patient (\textit{e.g.}, gender, age at the start of the admission period)
  \item Medications (\textit{e.g.}, numbers prescribed and administered)
  \item Diagnoses (\textit{e.g.}, presence or absence)
  \end{itemize}
These are included to establish whether some of these variables can be correlated with violence incidents.

  The resulting dataset consists of 4280 admission periods, corresponding to 2892 unique patients. The dataset is highly imbalanced, as a mere 425 admission periods have a violent outcome. In further sections, we will discuss how the imbalanced nature of the dataset affects the analysis.
  
  \section{Methodology}
  In this work, we address the problem of violence risk prediction as a document classification task, where EHR document features are combined with additional structured data, as explained in Sec.~\ref{sec:dataset}. For text normalization purposes, we perform a series of pre-processing steps outlined in Sec.~\ref{sec:text_normalization}. Then, for document representation purposes, we experiment with two alternative approaches---paragraph embeddings and LDA topic vectors---discussed in Sec.~\ref{sec:text_reps}. For the classification task, we experiment with SVM~\cite{svm} and random forest classification~\cite{breiman} (Sec.~\ref{sec:classifiers}). Finally, we discuss our choice of evaluation metrics in Sec.~\ref{sec:choice_of_metric}.

  \subsection{Text normalization}
  \label{sec:text_normalization}
  All notes are pre-processed by applying the following normalization steps:
  \begin{itemize}
  \item Converting all period notes to lowercase
  \item Removing special characters (\textit{e.g.}, \"{e} $\rightarrow$ e)
  \item Removing non-alphanumeric characters
  \item Tokenizing the texts using the NLTK Dutch word tokenizer~\cite{nltk_book}
  \item Removing stopwords using the default NLTK Dutch stopwords list
  \item Stemming using the NLTK Dutch snowball stemmer
  \item Removing periods
  \end{itemize}
  
  \subsection{Text representations}
  \label{sec:text_reps}
  The language used in clinical text is domain-specific, and the notes are rich in technical terms and spelling errors. Pre-trained paragraph embedding models do not necessarily yield useful representations. For this reason, we use the entire available set of 834834 de-identified clinical notes to train both the paragraph embedding model and the topic model. Only notes with at least 10 words each are used, to remove notes that contain no valuable information.
  \subsubsection{Paragraph embeddings} We use Doc2Vec~\cite{paragraph2vec} to convert texts to paragraph embeddings. The Doc2Vec training parameters are set to the default values in Gensim 3.8.1~\cite{gensim-doc}, with the exception of four parameters. Namely, we increase {\tt epochs} from 5 to 20 to improve the probability of convergence; we increase {\tt min\_count}---the minimum number of times a word has to appear in the corpus in order to be considered---from 5 to 20 to avoid including repeated mis-spellings of words~\cite{vincent6}; we increase {\tt vector\_size} from 100 to 300 to enrich the vectors while keeping the training time acceptable; and we decrease {\tt window}---the size of the context window--- from 5 to 2 to mitigate the effects of the lack of structure often present in EHR texts.

  \subsubsection{Topic modeling} As an alternative to paragraph embeddings we consider topic modeling. In topic modeling, you first analyse a sample of texts looking for collections of words that represent topics. You can then compute, for each text, to what degree it expresses each topic. This results in a vector of weights for each topic. A previous study using Latent Dirichlet Allocation (LDA) for topic modeling in the psychiatry domain~\cite{rumshisky} suggests that topic modeling can be used alternatively or in addition to text embeddings in classification problems. We use the LdaMallet~\cite{gensim-doc} implementation of LDA to train a topic model on a large number of clinical notes. In order to determine the optimal number of topics, we use the coherence model implemented in Gensim to compute the coherence metric~\cite{syed_coherence}. We find the optimal number to be 25. We use default values for the LdaMallet training parameters.

  \subsection{Classification methods}
  \label{sec:classifiers}
  Similarly to previous work~\cite{vincent5,vincent6}, we use Support Vector Machines (SVM)~\cite{svm}. Moreover, for interpretability purposes, we implement in this work random forest classification~\cite{breiman}. Random forests are ensemble models that are widely used in classification problems~\cite{ideal2019}. The {\tt scikit-learn} implementation of random forests outputs after training a list of the most relevant features used for classification. This can help us determine whether some of the features are more important than others when it comes to classifying positive and negative samples.

  We use two loops of 5-fold cross-validation for estimation of uncertainty and hyper-parameter tuning. In each iteration of the outer loop, the admission periods corresponding to 1/5 of the patients are kept as test data, and the remaining admission periods are used in the inner loop to perform a grid search for hyper-parameter tuning. The best classifier from the inner loop is applied to the test data, and the resulting classification metrics from each iteration of the outer loop are used to calculate a mean and a standard deviation for the metrics.

  We employ the SVC support-vector classifier provided by {\tt scikit-learn}, with default parameters except for the following: {\tt class\_weight} is set to `balanced' to account our imbalanced dataset; {\tt probability} is True to enable probability estimates for performance evaluation; the cost parameter {\tt C} and the kernel coefficient {\tt gamma} are determined by cross-validation. The ranges of values used are {\tt C} = \{$10^{-1}$, $10^{0}$, $10^{1}$\} and {\tt gamma} = \{$10^{-5}$, $10^{-4}$, $10^{-3}$, $10^{-2}$, $10^{-1}$, $10^{0}$\}. Both of these ranges were motivated in a previous study~\cite{vincent6}.

  For the random forest classifier, we use the {\tt scikit-learn} implementation, with default values for all the parameters except for the following: {\tt n\_estimators} is increased to 500 to prevent overfitting; {\tt class\_weight} is set to `balanced' to account for the imbalanced dataset; and {\tt min\_samples\_leaf}, {\tt max\_features} and {\tt criterion} are determined by cross-validation. Values for {\tt min\_samples\_leaf} are greater than the default value of 1, to prevent overfitting. For {\tt max\_features}, we consider the default value of `auto', which sets the maximum number of features per split to the square root of the number of features, and two smaller values, again in order to prevent overfitting. Finally, both split criteria available in {\tt scikit-learn} were considered (`gini' and `entropy'). These parameters are summarized in Tab.~\ref{tab:rf_params}.
  \begin{table}
    \centering
\caption{Parameters used for random forest classifier training. Parameters with multiple values are varied in a cross-validation search. The remaining parameters are set to the default {\tt scikit-learn} values.}\label{tab:rf_params}
\begin{tabular}{|l|l|l|}
\hline
Parameter &  Value/s & Method \\
\hline
{\tt min\_samples\_leaf} & \{3, 5, 10\} & Cross-validation \\
{\tt max\_features} & \{5.2, 8.7, `auto'\} & Cross-validation \\
{\tt criterion} & \{`gini', `entropy'\} & Cross-validation \\
\hline
{\tt n\_estimators} & 500 & Fixed \\
{\tt class\_weight} & `balanced' & Fixed \\
\hline
\end{tabular}
\end{table}

  \subsection{Evaluation metrics}
  \label{sec:choice_of_metric}
  
  Binary classifiers predict probabilities for input samples to belong to the positive class. When employing a binary classifier in practice, a threshold is chosen, and all samples with positive probabilities above that thresholds are considered positive \emph{predictions}. While testing the performance of a classifier, then, we can compare the actual \emph{conditions} with the predictions, and classify each sample as a \emph{true positive} (TP), \emph{true negative} (TN), \emph{false positive} (FP) or \emph{false negative} (FN) (see Tab.~\ref{tab:confusion_matrix}).
  \begin{table}
    \centering
  \caption{Definitions of the four conditions that a classified sample can belong to in binary classification. P and N stand for positive and negative, respectively. T and F stand for true and false, respectively.}
  \label{tab:confusion_matrix}
\begin{tabular}{|c|c|c|}
\hline
\multirow{2}{*}{Predictions} &  \multicolumn{2}{c|}{Conditions}  \\
\cline{2-3}
& P & N \\
\hline
P & TP & FP \\
\hline
N & FN & TN \\
\hline
\end{tabular}
\end{table}

Choosing an operating threshold in practice requires domain expert knowledge. \textit{E.g.}, if violence incidents have very high costs (human or economic), avoiding false negatives would be a priority; if, on the other hand, interventions are costly and cannot be afforded for most patients, avoiding false positives would be more important. Because the decision of the operating threshold is usually made only when the classifier will be put into practice, we report the performance of classifiers using metrics that are agnostic to the operating threshold.

The performance of classifiers is often reported in terms of the Receiver Operating Characteristic Area Under the Curve (ROC-AUC)~\cite{vincent5,vincent6}. The ROC is a plot of the true positive rate (TPR) as a function of the false positive rate (FPR), where TPR and FPR are defined as
\begin{equation}
  {\rm TPR} = \frac{{\rm TP}}{{\rm TP} + {\rm FN}};~
  {\rm FPR} = \frac{{\rm FP}}{{\rm TN} + {\rm FP}}
\end{equation}
In other words, the curve is constructed by choosing multiple classification thresholds, computing the quantities in Tab.~\ref{tab:confusion_matrix} for each threshold, then computing FPR and TPR. As you vary the classification threshold, you allow more or fewer positive predictions, so FPR and TPR both vary in the same direction. In a random classifier, FPR and TPR vary at the same rate, so the baseline ROC is a straight line between (0,0) and (1,1), and the baseline ROC-AUC is 0.5. The maximum ROC-AUC is 1, which represents perfect discrimination between TP and FP.

It has been previously noted~\cite{saito} that ROC-AUC is not a robust performance metric when dealing with imbalanced datasets such as ours. Because the dataset is highly imbalanced, the FPR can be misleadingly small, simply because the denominator includes all negative samples, and this artificially increases the ROC-AUC.
For this reason, we opt in this work to implement the area under the Precision-Recall curve (PR-AUC) evaluation measure~\cite{prcurve}.
The Precision-Recall curve is, as its name suggests, a plot of the precision of the classifier as a function of its recall, with precision and recall defined as:
  \begin{equation}
    {\rm Precision} = \frac{{\rm TP}}{{\rm TP} + {\rm FP}};~
    {\rm Recall} = \frac{{\rm TP}}{{\rm TP} + {\rm FN}}
  \end{equation}
  Note that neither of these quantities are directly dependent on TN, which is a desirable feature because we have an imbalanced dataset with a large number of negatives, and we are more interested in the few positives.

  To determine the baseline value for PR-AUC, note that, no matter what recall you get, the best precision you can get by guessing randomly is the real fraction of positive samples, $f_P$. Thus, the baseline PR-AUC is $f_P$, which in our case is 425/4280 = 0.10.
  
  Though we make no decision regarding the classification threshold, we believe that due to the nature of violence incidents it is more important to avoid FN than to avoid FP. Thus, a good metric to quantify the performance of a classifier in practice is $F_2$, given by:
  \begin{equation}
    F_\beta = (1+\beta^2)\cdot\frac{{\rm Precision}\cdot{\rm Recall}}{\beta^2\cdot{\rm Precision} + {\rm Recall}}
  \end{equation}
  with $\beta = 2$~\cite{fbeta}.

  In this work, we report our classifier performance in both PR-AUC and ROC-AUC, for comparison with previous work on similar datasets~\cite{vincent5,vincent6,rumshisky,suchting}. We also report $F_2^{\rm max}$, \textit{i.e.}, the value of $F_2$ at the classification threshold that maximizes $F_2$.

  \section{Experimental Results}
  \subsection{Classifier performance}
\begin{figure}
\centering \includegraphics[width=0.9\textwidth]{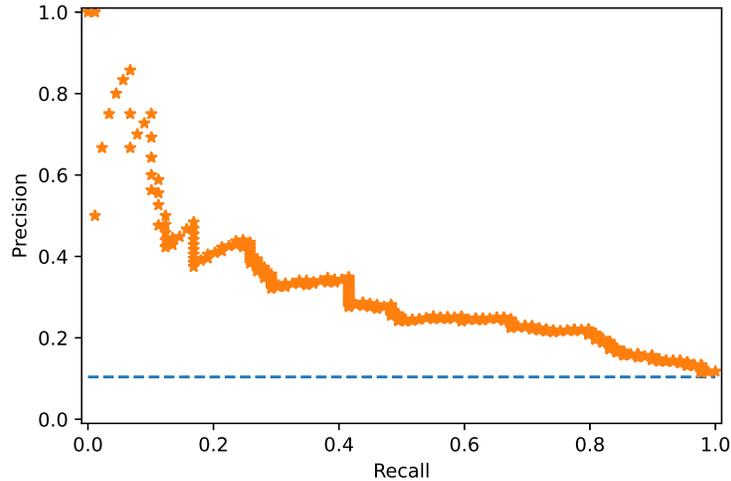}
\caption{Precision-Recall curve for one of the folds of the uncertainty-estimation loop during training of the SVM classifier. The PR-AUC is 0.33.}
\label{fig:prc}
\end{figure}

  Tab.~\ref{tab:results} reports the results of the analyses. All configurations gave results consistent with each other, as well as with previous work on a smaller dataset~\cite{vincent6}.
\begin{table}
  \centering
\caption{Classification metrics for various training configurations.}\label{tab:results}
\begin{tabular}{|l|l|l|l|l|l|l|}
\hline
LDA &  Embeddings & Structured vars & Estimator & PR-AUC & ROC-AUC & $F_2^{\rm max}$ \\
\hline
No & Yes & No & SVM & 0.321$\pm$0.067 & 0.792$\pm$0.011 & 0.519 \\
No & Yes & No & RF & 0.293$\pm$0.054 & 0.782$\pm$0.011 & 0.514 \\
No & Yes & Yes & RF & 0.299$\pm$0.056 & 0.782$\pm$0.011 & 0.515 \\
Yes & No & Yes & RF & 0.309$\pm$0.070 & 0.785$\pm$0.011 & 0.503 \\
Yes & Yes & Yes & RF & 0.304$\pm$0.058 & 0.792$\pm$0.011 & 0.517 \\
\hline
\end{tabular}
\end{table}
Fig.~\ref{fig:prc} shows the precision-recall curve for one of the folds of the outer uncertainty-estimation loop during the training of the SVM classifier.

  \subsection{Feature importance}
\begin{figure}
\centering \includegraphics[width=0.9\textwidth]{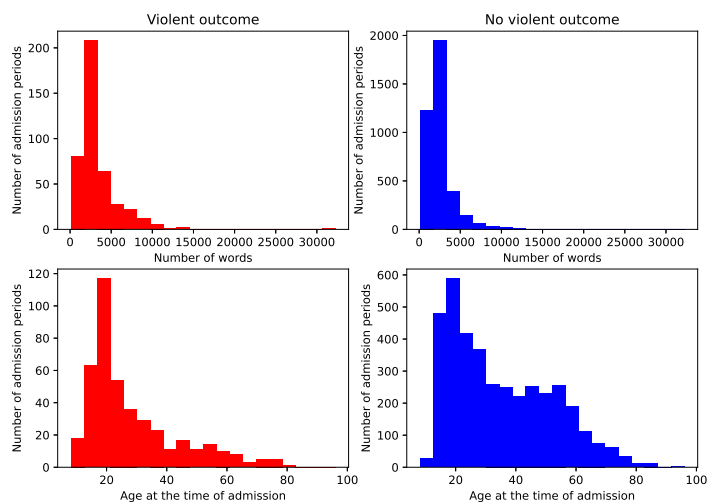}
\caption{Histograms of number of words per period note (top) and age at the beginning of the admission period (bottom) for violent (left, colored red) and non-violent (right, colored blue) patients.} \label{fig:feat_imp}
\end{figure}

  When using the random forest estimator, at each step in the outer cross-validation loop we stored the 10 most important features according to the best fit in the inner cross-validation loop. Gathering all the most important features together, we then studied both the 10 most repeated features and the 10 features with the highest total feature importance. These lists were reassuringly similar. The most repeated features were 5 of the text embedding features, plus the age at the beginning of the admission period ({\tt age\_admission}) and the number of words in the period note ({\tt num\_words}). The frequency distributions of these variables are shown, for both positive and negative samples, on Fig.~\ref{fig:feat_imp}.
As can be seen in the figures, the average violent patient is younger than the average non-violent patient, and the average period note about a violent patient is longer than the average period note about a non-violent patient.

\subsection{Inter-classifier agreement}
\begin{figure}
\centering \includegraphics[width=0.9\textwidth]{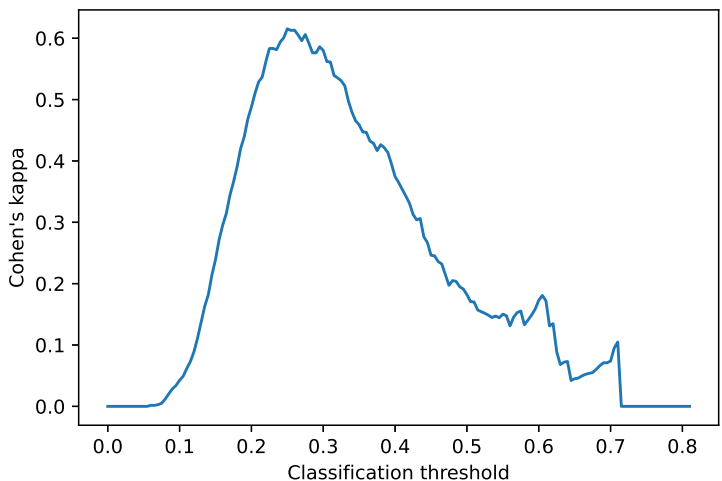}
\caption{Cohen's kappa score for inter-classifier reliability, comparing the LDA-topic-model-based classifier and the document-embeddings-based classifier, as a function of the classification threshold. 200 equidistant classification thresholds between 0 and 1 were used. The same thresholds were applied to both classifiers.}
\label{fig:cohenskappa}
\end{figure}

To better understand why paragraph embeddings and topic models gave similar classification metrics, we studied the inter-classifier agreement using Cohen's kappa~\cite{cohenskappa}. This metric quantifies how much two classifiers agree, taking into consideration the probability that they agree by chance. A value of Cohen's kappa equal to 0 means the agreement between the two classifiers is random, while a value of 1 means the agreement is perfect and non-random. We placed classification thresholds at multiple points between 0 and 1 (same threshold for both classifiers), and computed classification labels for each classifier for each threshold. Using those classification labels, we computed Cohen's kappa. The result is shown in Fig.~\ref{fig:cohenskappa}.

Next, for each classifier, we used the threshold that maximized the $F_2$ metric (see Sec.~\ref{sec:choice_of_metric}), and calculated classification labels for each classifier using those thresholds. We then computed Cohen's kappa using those classification labels, and obtained a value of:
\begin{equation}
  \kappa = 0.633\pm0.012,
\end{equation}
which is close to the maximum value reported in Fig.~\ref{fig:cohenskappa}.

According to the standard interpretation of Cohen's kappa, this value implies that the agreement between the two classifiers is better than random, though not very close to perfect agreement. This might indicate that the paragraph embedding representations and those based on LDA are capturing similar information from the notes.

\section{Conclusions}
We applied machine learning methods to Dutch clinical notes from the psychiatry department of the University Medical Center (UMC) in Utrecht, the Netherlands. We trained a model that predicts, based on the content of the notes, which patients are likely to be involved in a violence incident within their first 4 weeks of admission. The performance of our classifiers is assessed using the area under the precision-recall curve (PR-AUC), and its value is approximately 0.3, well above the baseline value of around 0.1; the ROC-AUC is approximately 0.8. The maximum $F_2$ score, which puts twice as much importance on recall as on precision, is approximately 0.5. 
Our results are competitive with a study based on structured variables that obtained ROC-AUC = 0.7801~\cite{suchting}.
These metrics show modest performance, and they indicate that further work is needed to extract all the meaningful information contained in the clinical notes.

The fact that only two of the structured variables included in our study---number of words and patient age---resulted in significant differentiation between the positive and negative classes further stresses that novel sophisticated methods are required. In particular, deep learning is a promising approach since the appearance of Dutch-language models~\cite{bertje}.

We have also, for the first time as far as we are aware, applied topic modeling to clinical notes in Dutch language for Violence Risk Assessment. We found that the performance of classifiers on numerical representations produced by topic models is comparable to the performance of similar classifiers on document embeddings.

We note that this approach does not replace domain expertise, since the input to our methods are notes written by nurses and psychiatrists. Rather, this is a complementary approach that builds on existing knowledge, while combining it with powerful modern statistical tools.

\bibliographystyle{splncs04}
\bibliography{thebibliography}

\begin{thebibliography}{10}
\providecommand{\url}[1]{\texttt{#1}}
\providecommand{\urlprefix}{URL }
\providecommand{\doi}[1]{https://doi.org/#1}

\bibitem{Abbeetal2016}
Abbe, A., Grouin, C., Zweigenbaum, P., Falissard, B.: Text mining applications
  in psychiatry: A systematic literature review. MPR  \textbf{25}(2),  86--100
  (2016)

\bibitem{AggarwalZhai2012}
Aggarwal, C.C., Zhai, C.: Mining Text Data. Springer, Boston, MA (2012)

\bibitem{nltk_book}
Bird, S., Klein, E., Loper, E.: Natural Language Processing with Python:
  Analyzing Text with the Natural Language Toolkit,
  \url{https://www.nltk.org/book/}. Last accessed 13 Jun 2020

\bibitem{breiman}
Breiman, L.: Random forests. Machine Learning  \textbf{45},  5--32 (2001)

\bibitem{cohenskappa}
Cohen, J.: A coefficient of agreement for nominal scales. Educational and
  Psychological Measurement  \textbf{20}(1),  37--46 (1960)

\bibitem{Conneretal2012}
Conner, K.R., et~al.: Mental disorder comorbidity and suicide among 2.96
  million men receiving care in the veterans health administration health
  system. J. Abnorm. Psychol  \textbf{122}(1),  256–--263 (2012)

\bibitem{Cooketal2016}
Cook, B.L., Progovac, A.M., Chen, P., Mullin, B., Hou, S., Baca-Garcia, E.:
  Novel use of natural language processing {(NLP)} to predict suicidal ideation
  and psychiatric symptoms in a text-based mental health intervention in
  {M}adrid. Comput Math Methods Med  (2016)

\bibitem{svm}
Cortes, C., Vapnik, V.: Support-vector networks. Machine Learning
  \textbf{20}(3),  273--297 (1995)

\bibitem{Friedmanetal1983}
Friedman, S., Margolis, R., David, O.J., Kesselman, M.: {Predicting psychiatric
  admission from an emergency room. Psychiatric, psychosocial, and
  methodological factors}. J. Nerv. Ment. Dis  \textbf{171}(3),  155--158
  (1983)

\bibitem{Huangetal2014}
Huang, S.H., LePendu, P., Iyer, S.V., Tai-Seale, M., Carrell, D., Shah, N.H.:
  Toward personalizing treatment for depression: predicting diagnosis and
  severity. JAMIA  \textbf{21}(6),  1069--1075. (2014)

\bibitem{inoue}
Inoue, M., Tsukano, K., Muraoka, M., Kaneko, F., Okamura, H.: Psychological
  impact of verbal abuse and violence by patients on nurses working in
  psychiatric departments. Psychiatry and Clinical Neurosciences  \textbf{60},
  29--36 (2006)

\bibitem{kim}
Kim, Y.K. (ed.): Frontiers in Psychiatry: Artificial Intelligence, Precision
  Medicine, and Other Paradigm Shifts. Springer, Singapore (2019)

\bibitem{paragraph2vec}
Le, Q., Mikolov, T.: Distributed representations of sentences and documents.
  In: ICML'14. pp. 1188--1196. PMLR, Beijing, China (2014)

\bibitem{vanleeuwen}
van Leeuwen, M., Harte, J.: Violence against mental health care professionals:
  prevalence, nature and consequences. J. Forensic Psychiatry Psychol.
  \textbf{28}(5),  581--598 (2017)

\bibitem{Lyonsetal1997}
Lyons, J.S., Stutesman, J., Neme, J., Vessey, J.T., O'Mahoney, M.T., Camper,
  H.J.: Predicting psychiatric emergency admissions and hospital outcome.
  Medical care  \textbf{35}(8),  792–800 (1997)

\bibitem{prcurve}
Manning, C., Raghavan, P., Sch\"utze, H.: Introduction to Information
  Retrieval. Cambridge University Press, USA (2008)

\bibitem{vincent5}
Menger, V., Scheepers, F., Spruit, M.: Comparing deep learning and classical
  machine learning approaches for predicting inpatient violence incidents from
  clinical text. Applied Sciences  \textbf{8}(6), ~981 (2018)

\bibitem{deduce}
Menger, V., Scheepers, F., van Wijk, L., Spruit, M.: {DEDUCE: A pattern
  matching method for automatic de-identification of dutch medical text}.
  Telematics and Informatics  \textbf{35}(4),  727--736 (2018)

\bibitem{vincent6}
Menger, V., Spruit, M., van Est, R., Nap, E., Scheepers, F.: Machine learning
  approach to inpatient violence risk assessment using routinely collected
  clinical notes in electronic health records. JAMA Network Open
  \textbf{2}(7),  e196709 (2019)

\bibitem{nijman}
Nijman, H., Bowers, L., Oud, N., Jansen, G.: Psychiatric nurses' experiences
  with inpatient aggression. Aggressive Behavior  \textbf{31}(3),  217--227
  (2005)

\bibitem{Olfsonetal2011}
Olfson, M., Ascher-Svanum, H., Faries, D.E., Marcus, S.C.: Predicting
  psychiatric hospital admission among adults with schizophrenia. Psychiatric
  services  \textbf{62}(10),  1138–--1145 (2011)

\bibitem{Poulinetal2014}
Poulin, C., et~al.: Predicting the risk of suicide by analyzing the text of
  clinical notes. PLOS ONE  \textbf{9}(3) (2014)

\bibitem{gensim-doc}
{\v R}eh{\r u}{\v r}ek, R., Sojka, P.: Software framework for topic modelling
  with large corpora. In: Proceedings of LREC 2010 Workshop on New Challenges
  for NLP Frameworks. pp. 45--50. University of Malta, Valletta, Malta (2010)

\bibitem{Rijoetal2015}
Rijo, R., Martinho, R., Pereira, L., Silva, C.: Text mining applied to
  electronic medical records: A literature review. Int. J. E-Health Med.
  Commun.  \textbf{6}(3),  1--18 (2015)

\bibitem{fbeta}
van Rijsbergen, C.J.: Information Retrieval. 2nd edn. Butterworth-Heinemann
  (1979)

\bibitem{rumshisky}
Rumshisky, A., et~al.: Predicting early psychiatric readmission with natural
  language processing of narrative discharge summaries. Transl Psychiatry
  \textbf{6}, ~e921 (2016)

\bibitem{saito}
Saito, T., Rehmsmeier, M.: {The precision-recall plot is more informative than
  the ROC plot when evaluating binary classifiers on imbalanced datasets}. PLOS
  ONE  \textbf{10}(3),  e0118432 (2015)

\bibitem{singh}
Singh, J., et~al.: International perspectives on the practical application of
  violence risk assessment: A global survey of 44 countries. International
  Journal of Forensic Mental Health  \textbf{13}(3),  193--206 (2014)

\bibitem{suchting}
Suchting, R., Green, C.E., Glazier, S.M., Lane, S.D.: A data science approach
  to predicting patient aggressive events in a psychiatric hospital. Psychiatry
  Research  \textbf{268},  217--222 (2018)

\bibitem{syed_coherence}
Syed, S., Spruit, M.R.: {Full-Text or Abstract? Examining Topic Coherence
  Scores Using Latent Dirichlet Allocation}. In: DSAA2017. pp. 165--174 (2017)

\bibitem{bertje}
de~Vries, W., van Cranenburgh, A., Bisazza, A., Caselli, T., van Noord, G.,
  Nissim, M.: {BERTje: A Dutch BERT Model},
  \url{https://arxiv.org/abs/1912.09582}

\bibitem{Wangetal2018}
Wang, Y., et~al.: Clinical information extraction applications: A literature
  review. Journal of Biomedical Informatics  \textbf{77},  34--49 (2018)

\bibitem{ideal2019}
Yin, H., Camacho, D., Tino, P., Tallón-Ballesteros, A., Menezes, R.,
  Allmendinger, R. (eds.): IDEAL 2019, vol. 11872. Springer (2019)

\end{thebibliography}

%  \bibitem{ref_article1}
%  Author, F.: Article title. Journal \textbf{2}(5), 99--110 (2016)
%Author, F., Author, S.: Title of a proceedings paper. In: Editor,
%F., Editor, S. (eds.) CONFERENCE 2016, LNCS, vol. 9999, pp. 1--13.
%Springer, Heidelberg (2016). \doi{10.10007/1234567890}
%Author, F., Author, S., Author, T.: Book title. 2nd edn. Publisher,
%Author, A.-B.: Contribution title. In: 9th International Proceedings
%on Proceedings, pp. 1--2. Publisher, Location (2010)
%LNCS Homepage, \url{http://www.springer.com/lncs}. Last accessed 4
\end{document}